\documentclass[10pt,twocolumn,letterpaper]{article}

\usepackage{cvpr}              % To produce the CAMERA-READY version
\usepackage{appendix}
\usepackage{graphicx}
\usepackage{amsmath}
\usepackage{amssymb}
\usepackage{booktabs}
\usepackage{tikz}
\usepackage{comment}
\usepackage{color}
\usepackage{multirow}
\usepackage{epsfig}
\usepackage{float}
\usepackage[misc]{ifsym}
\usepackage{caption}
\usepackage{array}
\usepackage{url}
\usepackage{cuted}

\usepackage[capitalize]{cleveref}
\crefname{section}{Sec.}{Secs.}
\Crefname{section}{Section}{Sections}
\Crefname{table}{Table}{Tables}
\crefname{table}{Tab.}{Tabs.}

\def\confName{CVPR}
\def\confYear{2022}

\begin{document}
\captionsetup[figure]{labelfont={bf},labelformat={default},labelsep=period,name={Fig.}}
\title{GP-NAS-ensemble: a model for NAS Performance Prediction}

\author{Kunlong Chen$^{1}$ \quad Liu Yang$^{2}$ \quad Yitian Chen$^{3}$ \quad Kunjin Chen$^{4}$ \quad  Yidan Xu$^{1}$ \quad Lujun Li$^{5}$ \\
{\normalsize $^{1}$Meituan \quad $^{2}$Tencent \quad $^{3}$Bigo Technology \quad $^{4}$Alibaba Group \quad $^{5}$Chinese Academy of Sciences}\\
{\tt\small $^{1}$\{chenkunlong, xuyidan02\}@meituan.com, $^{2}$arielwillow@163.com,} \\ {\tt\small $^{3}$yitiansky@gmail.com $^{4}$kunjin.ckj@alibaba-inc.com, $^{6}$lilujunai@gmail.com}
}

\maketitle

%%%%%%%%% ABSTRACT
\begin{abstract}
It is of great significance to estimate the performance of a given model architecture without training in the application of Neural Architecture Search (NAS) as it may take a lot of time to evaluate the performance of an architecture. In this paper, a novel NAS framework called GP-NAS-ensemble is proposed to predict the performance of a neural network architecture with a small training dataset. We make several improvements on the GP-NAS model to make it share the advantage of ensemble learning methods. Our method ranks second in the CVPR2022 second lightweight NAS challenge performance prediction track.
\end{abstract}

%%%%%%%%% BODY TEXT

\section{Introduction}
With the development of Neural Architecture Search (NAS) techniques,  it becomes popular to design deep neural network architectures automatically \cite{baidua}. It is a critical part to design a performance predictor to analyze the relationship between model architecture and its accuracy evaluated on certain tasks for NAS algorithms, as training a large net can be fairly expensive \cite{zhang2021cascade}. 

GP-NAS-ensemble model is proposed as a novel predictor in this paper. Based on GP-NAS model \cite{li2020gp}, we make several improvements to make it more accurate and robust. The validity of our method is verified in the Performance Prediction Track of CVPR2022 Second lightweight NAS challenge.

\subsection{Literature Review}
\noindent\textbf{General NAS.} 
In a wide range of computer vision tasks\cite{xu2019lightweightnet, gu2018recent, arun2019convolutional,ma2018shufflenet,yousef2020accurate,li2021nas,lishadow,li2022norm,li2020explicit,li2022SFF,li2022tf,li2022self,nas2}, manually constructed neural networks have had great success. Artificial designs are usually thought to be suboptimal. Both academia and industry have recently been more interested in neural architecture search (NAS). Early efforts made use of reinforcement learning \cite{DBLP:conf/cvpr/TanCPVSHL19, DBLP:conf/cvpr/ZhongYWSL18,  DBLP:conf/iclr/ZophL17, DBLP:conf/cvpr/ZophVSL18} and evolutionary algorithms\cite{DBLP:conf/aaai/RealAHL19, DBLP:conf/icml/RealMSSSTLK17, DBLP:conf/icml/SuganumaOO18, DBLP:conf/iccv/XieY17, DBLP:journals/corr/abs-1810-03522} and discovered several high-cost and high-performance designs.
Later work aims to lower the cost of searching while increasing performance, which can be divided into three categories: one-shot NAS \cite{guo2019single, zhang2020one, dong2019one,chu2019fairnas}, gradient-based approaches\cite{liu2018darts,chang2020data, wang2020you}, and predictor NAS, which differ in the network architecture modeling process. One-Shot NAS \cite{chu2019fairnas} firstly trains an over-parameterized supernet, then searches a discrete search space that includes numerous candidate models. The sample strategy is important during the training stage, since it determines how to train an effective supernet for performance estimation.  Gradient-based methods incorporate the architecture parameters for each operator and use backpropagation to jointly optimize them and the weights of the network.

\noindent\textbf{Predictor NAS.} 
Predictor NAS approaches attempt to accurately and effectively forecast the performance of a particular neural network. These procedures learns the accuracy predictor by sampling pairs of architectures and their accuracies. 
Some works~\cite{DBLP:conf/iclr/BakerGRN18, DBLP:conf/iclr/KleinFSH17, DBLP:conf/ijcai/DomhanSH15,liang} extend this line of thought by training a predictor to extrapolate the NAS learning curves.  The regression problem~\cite{DBLP:conf/eccv/WenLCLBK20} or ranking~\cite{DBLP:conf/eccv/NingZZWY20, DBLP:conf/cvpr/Xu00TJX021} can be used to describe the goal of fitting the predictor \cite{DBLP:journals/corr/abs-2003-12857, DBLP:conf/nips/LuoTQCL18, DBLP:conf/nips/DudziakCALKL20, DBLP:journals/corr/abs-2007-04785}. Moreover, feature representation and sampling methods are crucial for search performance. \cite{DBLP:conf/nips/LuoTQCL18} uses acyclic graphs in a continuous space of potential embeddings along with performance predictors. \cite{DBLP:conf/cvpr/Wang0GC21} improves the two-stage NAS with Pareto-aware sampling strategies. \cite{DBLP:conf/nips/ShiPXLKZ20} uses Bayesian  regression as a proxy model to select candidates, and \cite{wu2021weak} replaces a strong predictor with a set of weaker predictors.

\noindent\textbf{Multi-task NAS.} Multi-task learning refers to different tasks sharing part of the network backbone or weights. These tasks are often able to learn from each other to achieve better performance and training efficiency. Recent neural network architecture search methods and benchmarks~\cite{DBLP:conf/cvpr/DuanCXCLZL21, DBLP:journals/corr/abs-2110-05668,liang} for multi-task and cross-task have attracted a lot of attention from the community. Despite being underappreciated in comparison to single-task NAS, there are still several excellent algorithms. \cite{DBLP:conf/acl/PasunuruB19} uses continuous learning to find a single cell structure that can generalize well to unknown tasks via multi-task architecture search based on the weight sharing technique. \cite{DBLP:conf/accv/LiuLPZTS20} used gradient-based NAS to find the best cell structure for a variety of autonomous driving tasks. \cite{DBLP:conf/iclr/LeeHH21} proposed to construct graphs from datasets in a meta-learning approach to make the methods generalize well across numerous tasks.

\section{Proposed Method}
In this paper, we introduce the \textit{GP-NAS-ensemble} model \cite{li2020gp}, which ranked 2nd in the CVPR 2022 NAS competition: performance estimation track. Based on the GP-NAS model, we aim at establishing a model with better performance using ensemble learning technique. 

\subsection{GP-NAS}
GP-NAS is a powerful method to predict the performance of a neural network given its architecture, especially when the size of training data is small \cite{li2020gp}. To be more specific, the GP-NAS model uses the Gaussian process regression model to predict the accuracy of a neural network model under the assumption that the joint distribution between the training observations $\mathbf{y}$ and the test function values $f(\boldsymbol{X}_{*}) = \boldsymbol{f}_{*}$ is \cite{rasmussen2003gaussian}

\begin{strip}
\begin{equation}
p\left(\boldsymbol{y}_{,} \boldsymbol{f}_{*} \mid \boldsymbol{X}, \boldsymbol{X}_{*}\right) = \mathcal{N}\left(\left[\begin{array}{c}
m(\boldsymbol{X}) \\
m\left(\boldsymbol{X}_{*}\right)
\end{array}\right],\left[\begin{array}{cc}
\mathbf{K}+\sigma_{n}^{2} \mathbf{I} & k\left(\boldsymbol{X}^{\prime}, \boldsymbol{X}_{*}\right) \\
k\left(\boldsymbol{X}_{*}, \boldsymbol{X}\right) & k\left(\boldsymbol{X}_{*}, \boldsymbol{X}_{*}\right).
\end{array}\right]\right)
\end{equation}\end{strip}

We then have

\begin{strip}
\begin{equation}
\begin{split}
p\left(\boldsymbol{f}_{*} \mid \boldsymbol{X}, \boldsymbol{y}, \boldsymbol{X}_{*}\right) &= \mathcal{N}\left(\mathbb{E}\left[\boldsymbol{f}_{*} \mid \boldsymbol{X}, \boldsymbol{y}, \boldsymbol{X}_{*}\right], \mathbb{V}\left[\boldsymbol{f}_{*} \mid \boldsymbol{X}, \boldsymbol{y}, \boldsymbol{X}_{*}\right]\right), \\
\mathbb{E}\left[\boldsymbol{f}_{*} \mid \boldsymbol{X}, \boldsymbol{y}, \boldsymbol{X}_{*}\right] &= m_{\text {post }}\left(\boldsymbol{X}_{*}\right)=\underbrace{m\left(\boldsymbol{X}_{*}\right)}_{\text {prior mean }}+\underbrace{k\left(\boldsymbol{X}_{*}, \boldsymbol{X}\right)\left(\boldsymbol{K}+\sigma_{n}^{2} \boldsymbol{I}\right)^{-1}}_{\text {“Kalman gain” }} \underbrace{(\boldsymbol{y}-m(\boldsymbol{X}))}_{\text {error }},
\end{split}
\end{equation} 
\end{strip}
\noindent where $\mathbf{K}$ is the Gram matrix, which contains the kernel functions evaluating on all pairs of data points, $\sigma_n^2$ is the noise variance of observations.

Simply put, the Gaussian process model uses the prior mean as an initial guess for the target value. The guess is then corrected by a mechanism similar to Kalman filter. It is therefore important to note that, the cores of the GP-NAS method are two parts: (1) the estimation of the prior mean; (2) the specific formula of the kernel function. In the original implementation of the algorithm, the linear regression method is used to estimate the prior mean with a variant of the radial basis function (RBF) kernel:
\begin{equation}
\begin{split}
        m(\mathbf{x}) &= \mathbf{w} \cdot \mathbf{x}, \\
        k_{rbf}^s(\mathbf{x}_1, \mathbf{x}_2) &= \exp[-\sqrt{ \vert \vert (\mathbf{x}_1 - \mathbf{x}_2) \vert \vert} / l],
        \end{split}
\end{equation}
where $\mathbf{w}$ is the coefficient vector of the linear regression model and $l$ is the length scale of the kernel.

\subsection{Feature Engineering}
In the original dataset, an ordinal encoder is used to represent each architecture. For example, 
the search space of features such as depth of network, number of heads of each layer, and MLP ratio of each layer are $\{10, 11, 12\}$. They are encoded as $\{1, 2, 3\}$, respectively, in the feature space.

One-hot encoding is a sparse way of representing data in a binary string in which only a single bit can be 1, while all others are 0 \cite{onehot}. This method is considered as a popular way to deal with catogorical features in the machine learning community. However, the disadvantage of one-hot encoding is that it does not contain the information of the `similarity'' of two data points. For instance, the distance between ``3'' and ``1'' should be larger than the distance between ``2'' and ``1''. In order to solve this problem, we also use the two-hot encoding method to represent the similarity in the feature space. An example of applying one-hot encoding and two-hot encoding methods is shown in Table~\ref{tab1}.

\begin{table}
  \centering
  \begin{tabular}{l c c}
    \toprule
    original feature & one-hot encoding & two-hot encoding\\
    \midrule
    0 & $[0, 0, 0, 0]$ & $[0, 0, 0, 0]$\\
    1 & $[1, 0, 0, 0]$ & $[1, 1, 0, 0]$\\
    2 & $[0, 1, 0, 0]$ & $[0, 1, 1, 0]$\\
    3 & $[0, 0, 1, 0]$ & $[0, 0, 1, 1]$\\
    \bottomrule
  \end{tabular}
  \caption{The feature engineering methods used in our framework.}
  \label{tab1}
\end{table}

\begin{figure*}
  \centering
  \includegraphics[width=14cm]{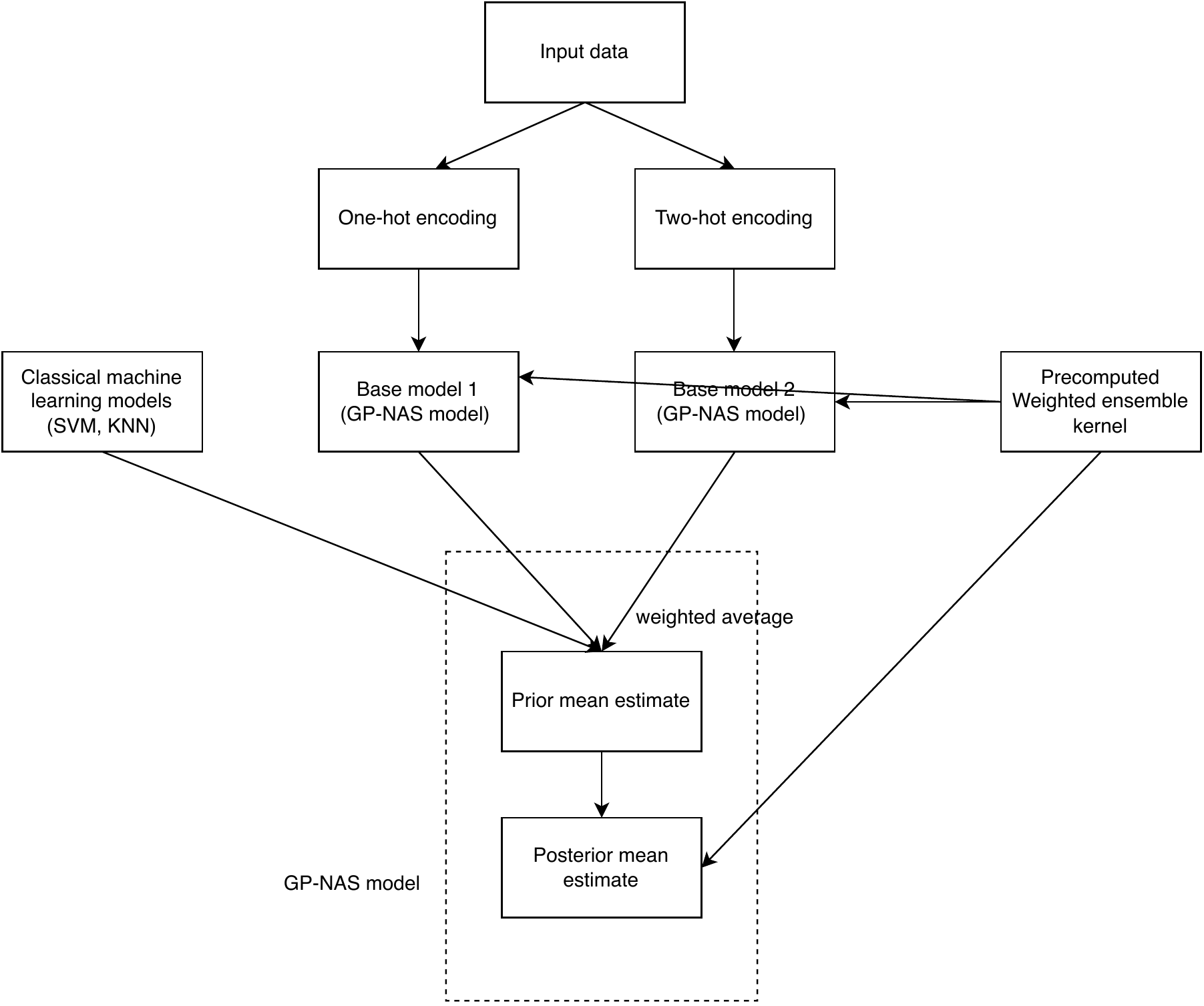}
  \caption{An illustration of the proposed GP-NAS-ensemble model.}
  \label{fig:short}
\end{figure*}

\begin{table*}
  \centering
  \begin{tabular}{l c c c c}
    \toprule
    Task & kernel length & weighted kernel ratio & Label transformation method & set of base learner\\
    \midrule
    task 0 & 22 & (0.18, 0.82) &  Normal distribution & GP-NAS, KNN\\
    task 1 & 28 & (0.62, 0.38) &  Left-skewed Normal distribution & GP-NAS, SVR\\
    task 2 & 24 & (0.02, 0.98) &  Left-skewed Normal distribution & GP-NAS, SVR\\
    task 3 & 25 & (0.6, 0.4) &  Normal distribution & GP-NAS, SVR\\
    task 4 & 22 & (0.7, 0.3) &  Left-skewed Normal distribution & GP-NAS, SVR\\
    task 5 & 22 & (0.3, 0.7) &  Normal distribution & GP-NAS, SVR\\
    task 6 & 22 &  - &  Normal distribution & GP-NAS\\
    task 7 & 22 & (0.3, 0.7) &  Normal distribution & GP-NAS, SVR\\
    \bottomrule
  \end{tabular}
  \caption{The configurations for each task.}
  \label{tab2}
\end{table*}

\begin{table}
  \centering
  \begin{tabular}{l c }
    \toprule
    Modifications & score (public)\\
    \midrule
    GP-NAS & $~0.668$\\
    + feature engineering & $~0.787$\\
    + label transformation & $0.796$ \\
    + ensemble learning & $0.798$ \\
    + weighted ensemble kernel & $0.800$ \\
    \bottomrule
  \end{tabular}
  \caption{The model performance on the public leaderboard.}
  \label{tab3}
\end{table}

\subsection{Label transformation}
The dataset does not provide us with the prediction accuracy of each architecture on test dataset. Instead, we only have access to the relative rank of each data point. We believe that it is more natural to predict the accuracy of each model architecture instead of predicting ranks through several trials. Since we do not have enough domain knowledge on this field, we need to guess the probability distribution of model performances and then assign each model with a score by sampling from the proposed distribution. There are two ways to guess the distribution of the accuracy: (1) they follow a normal distribution, (2) they follow a left-skewed bell-shaped distribution. We decide the distribution by conducting trials on the public leaderboard.

\subsection{Weighted ensemble kernel function}
The idea behind the weighted kernel function is that in each task, we should focus on different parts of the feature set. The weighted kernel function is 
\begin{equation}
    k_w(\mathbf{x}_1, \mathbf{x}_2) = \exp[-\sqrt{ (\mathbf{x}_1 - \mathbf{x}_2)^T I_w (\mathbf{x}_1 - \mathbf{x}_2)} / l],
\end{equation}
where $I_w$ is a diagonal matrix. In order to estimate $I_w$ for each task, we use the Bayesian optimization method to maximize the Kendall rank correlation coefficient on the training dataset.
Considering the fact that only a small amount of data is available, it may be prone to over-fitting by applying Bayesian optimization method directly. We know that if $k_1, k_2$ are valid kernels, $k_1 + k_2$ is still a valid kernel \cite{rasmussen2003gaussian}. We thus proposed a new kernel $k_w^e$, which is the weighted sum of $k_w$ and $k_{rbf}^s$:
\begin{equation}
    k_w^e(\mathbf{x}_1 , \mathbf{x}_2) = \beta_1 k_{rbf}^s(\mathbf{x}_1 , \mathbf{x}_2) + \beta_2 k_w(\mathbf{x}_1 , \mathbf{x}_2),
\end{equation}
where the values of $\beta_1$ and $\beta_2$ are selected by their performances on the public leaderboard.

\subsection{GP-NAS-ensemble model}
In this section, we briefly introduce the structure of our proposed GP-NAS-ensemble model. It adopts the modifications described in previous sections. The basic scheme of the GP-NAS-ensemble model is shown in Figure 1, which contains the following steps:
\begin{enumerate}
    \item Weighted ensemble kernel computing. We use the Bayesian optimization method to estimate the most suitable weighted ensemble kernel function for each task. 
    \item Base model establishment. Two GP-NAS models are built as base models. The difference lies in that we feed the data with one-hot encoding to the first model and we feed the data with two-hot encoding to the second one. In addition, several classical supervised learning methods are also considered as parts of our base-model set since we need to enhance the diversity between base models.
    \item Base model training. We train each base model separately on the training dataset.
    \item Ensemble model establishment. A GP-NAS-ensemble model can be built on top of the base models mentioned above. To be more specific, we predict the prior mean of a given test data by averaging the output of each base model. The estimation of the posterior mean is the same as the basic GP-NAS model.
\end{enumerate}

\section{Experiments}
The configurations of our final submission except for task 6 are shown in Table~\ref{tab2}. Specifically, we observe that the above-mentioned weighted ensemble kernel doesn't not increase the score of task 6. Therefore, we use nine-hot encoding method for the feature engineering step of task 6. The configurations are tuned based on scores on the public leaderboard.

In Table~\ref{tab3}, we show the results of an ablation study which compares the proposed model with the original GP-NAS model. The GP-NAS model can achieve about $0.668$ on public leaderboard. If we transform input features with one-hot encoding or two-hot encoding, the score is increased to about $0.787$. After label transformation with normal distribution, we obtain a score about $0.796$. When all modifications are applied, we get a final score close to $0.800$. 

\section{Conclusion}
Unlike most other competitors of this competition who use deep-learning method or other classical supervised machine learning to achieve a high score, we fully explored the potential of GP-NAS model with only small modifications on the model architecture and the feature engineering pipeline. The score of the proposed method on the public leaderboard increased from $0.668$ to about $0.800$.

%\vspace{-0.5cm}
% \paragraph{\textit{Limitations:}}Our neural rendering network still has to learn the mapping from paired underwater datasets, which limits the performance of our method. We hope to further explore the neural rendering by unsupervised learning in the future. 
{\small
\bibliographystyle{ieee_fullname}
\bibliography{egbib}
}

\newpage

\begin{appendices}

\end{appendices}
%%%%%%%%% REFERENCES

\end{document}